\documentclass{template/acm_proc_article-sp}
\usepackage{amsfonts}
\usepackage{verbatim}
\usepackage{multirow}
\usepackage{color}
\usepackage{graphicx}
\usepackage{graphics}
\usepackage{amssymb}
\usepackage{amsmath}
\usepackage[T1]{fontenc}

\begin{document}

\title{Bus Travel Time Predictions Using Additive Models}
\numberofauthors{3}
\author{
\alignauthor Matth\'ias Korm\'aksson\\
       \affaddr{IBM Research -- Brazil}\\
       \email{matkorm@br.ibm.com}
\and
\alignauthor Luciano Barbosa\\
       \affaddr{IBM Research -- Brazil}\\
       \email{lucianoa@br.ibm.com}
\and
\alignauthor Marcos R. Vieira\\
       \affaddr{IBM Research -- Brazil}\\
       \email{mvieira@br.ibm.com}
\and
\alignauthor Bianca Zadrozny\\
       \affaddr{IBM Research -- Brazil}\\
       \email{biancaz@br.ibm.com}
}

\maketitle

\begin{abstract}
Many factors can affect the predictability of public bus services such as traffic, weather and local events. Other aspects, such as day of week or hour of day, may influence bus travel times as well, either directly or in conjunction with other variables. However, the exact nature of such relationships between travel times and predictor variables is, in most situations, not known. In this paper we develop a framework that allows for flexible modeling of bus travel times through the use of Additive Models. In particular, we model travel times as a sum of linear as well as nonlinear terms that are modeled as smooth functions of predictor variables. 
The proposed class of models provides a principled statistical framework that is highly flexible in terms of model building. The experimental results demonstrate uniformly superior performance of our best model as compared to previous prediction methods when applied to a very large GPS data set obtained from buses operating in the city of Rio de Janeiro.
\end{abstract}

\section{Introduction}\label{sec:intro}

In this paper we are concerned with the problem of predicting bus travel/arrival times using GPS data from public buses. The main challenge in performing this task arises from the fact that GPS data only provide snapshots of bus locations at predefined (or in some cases irregular) time stamps. The observed GPS coordinates are therefore necessarily irregular in space as signal transmissions are not controlled with respect to bus locations. The difficulty of the problem is further increased when difference between time stamps is large.

The raw GPS data permit us to study the relationship between bus movements in time and space. However, other factors such as day of week, hour of day, and current traffic conditions may also influence travel times in some systematic way. The exact nature of such relationships between travel times and predictor variables is usually not known. Therefore, these factors need to be incorporated into prediciton algorithms either indirectly through binned analyses or through direct modeling.

We propose to model travel times using Additive Models~\cite{Hastie,Wood}, which provide a principled statistical framework for arrival time predictions. In particular, we model cumulative travel time as a smooth function of route location and further allow this functional relationship to vary smoothly across (clock) time. We also construct features that may seemlessly be incorporated into the Additive Model, either as direct main effects or interaction effects in conjunction with other variables.

Previous approaches have used a mixture of statistical and machine learning algorithms for predicting bus travel times. \cite{sun2007predicting,tiesyte2008similarity,lee2012http,sinn2012predicting} based predictions of future travel times on historical averages, either through binned analysis, e.g., with respect to hour of day, or by taking averages over similar past trips. \cite{williams2003,vanajakshi2009travel,shalaby2004prediction,chien2003} used Kalman filter or time series models to predict future travel times under the assumption of a direct relationship with previous travel times. The above approaches lack the ability to incorporate other features into the prediction algorithms in a model based manner.

As an alternative regression models provide a simple and highly interpretable framework for modeling travel time as a function of several features. However, \cite{sinn2012predicting,jeong2004} all demonstrated that the above models lack the flexibility to deal with nonlinear features so often present in these types of data. Artificial Neural Network (ANN) models and Support Vector Regression (SVR) models address this problem in a principled manner and have gained recent popularity in predicting bus arrival times because of their ability to deal with complex and nonlinear relationships between variables~\cite{chen2004dynamic,chien2002dynamic,wu2004travel,bin@jits2006}. However, these methods suffer from slow learning process \cite{altinkaya,hagan,bin@jits2006} and are difficult to interpret and implement unlike regression models.

A recurring problem in the above approaches is that they assume knowledge of travel times between fixed locations in space, in particular bus stops. Often times these data are available (e.g., Automatic Passenger Count (APC) data~\cite{shalaby2004prediction,chen2004dynamic}) and provide information about exact arrival, departure, and dwelling times at specified bus stops. In the absence of such data, interpolation is performed to infer these times at the route's bus stops~\cite{lee2012http,sinn2012predicting}. This is reasonable when difference between time stamps is small, say 20 seconds, but can lead to larger errors when difference is larger, say few minutes. Another problem arises for methods that account for temporal effects (e.g., Kalman filters) due to discretization made in the time dimension. This is again reasonable in the presence of high volumes of data, but may be problematic if data is sparse with irregularities in the time dimension.

The main advantage of Additive Models in this context is their ease of interpretability and flexibility in modeling complex non-linear relationships. Factors that are known (or suspected) to affect traffic may be included in the model as traditional linear features, smooth functional effects, or interactions thereof. Additive Models do not require any discretization or interpolated observations, but rather are capable of handling directly the raw observed data. The only interpolation that applies is made when inferring the departure time from origin. However, a critical feature of our proposed solution is the inclusion of a (corrective) random intercept in the model that attempts to correct for this interpolation step thus redefining time zero for each bus. Experimental results show that the random intercept model uniformly dominates all other methods in all prediction scenarios.  

To the best of our knowledge our proposed solution is the first method that: \textbf{(1)} models bus travel times directly using raw irregular GPS data; \textbf{(2)} models spatial and temporal effects through smooth functions thus avoiding any discretization; and \textbf{(3)} allows for flexible incorporation of additional traffic related features in a model based manner. The last point is an important one as it implies that our proposed framework may be used as a development framework for building more accurate travel time models through the incorporation of additional (perhaps city dependent) features.

The remainder of the paper is organized as follows: Section~\ref{sec:prediction} provides a summary of the motivating GPS data; background on additive models is provided in Section~\ref{sec:background}; the proposed framework for predicting bus travel times is detailed in Section~\ref{sec:modeling}; experimental evaluation is provided in Section~\ref{sec:experiments};
related work is described in Section~\ref{sec:related}; and Section~\ref{sec:conclusion} concludes the paper.

\section{Preliminaries} \label{sec:prediction}

In this section we start by describing the motivating data. We then explain how the data are normalized and introduce mathematical notation. 

\begin{figure}[!t]
	\centering
	\includegraphics[width=\linewidth]{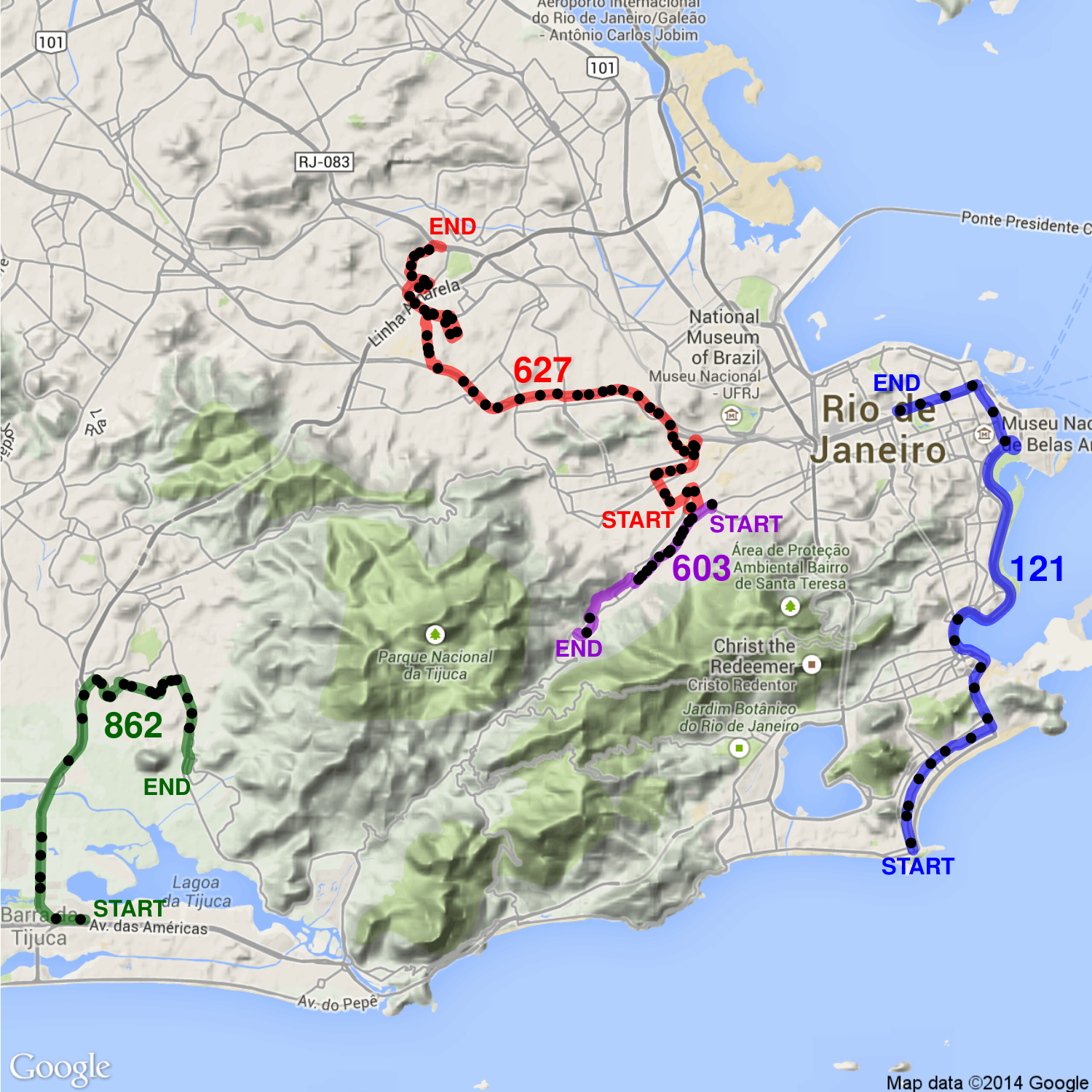}
	\caption{Map of the piecewise linear representations of the four routes analyzed in this paper. Route 603 (purple), 627 (red), 862 (green) and 121 (blue). Bus stops are marked by black points. }
	\label{fig:Routeggmap_all}
\end{figure}

\subsection{Motivating Data}

The motivating data consist of GPS measurements collected from public buses in the city of Rio de Janeiro, Brazil, during the time period from September 26, 2013 to January 9, 2014. The complete data set contains information about more than $400$ bus routes and $9000$ buses. Each GPS data point contains information about the position of the bus (longitude, latitude), date and time stamp, bus ID, and route ID. In total there are more than 100 million location entries for the time period of this study. The time between consecutive GPS measurements ranges from anywhere under a minute to over $10$ minutes, with an average of $\approx 4$ minutes. A sequence of GPS coordinates of a given bus is called a \emph{space-time trajectory} and provides information about bus movement in space and time.

\begin{figure}[!t]
	\centering
	\includegraphics[width=\linewidth]{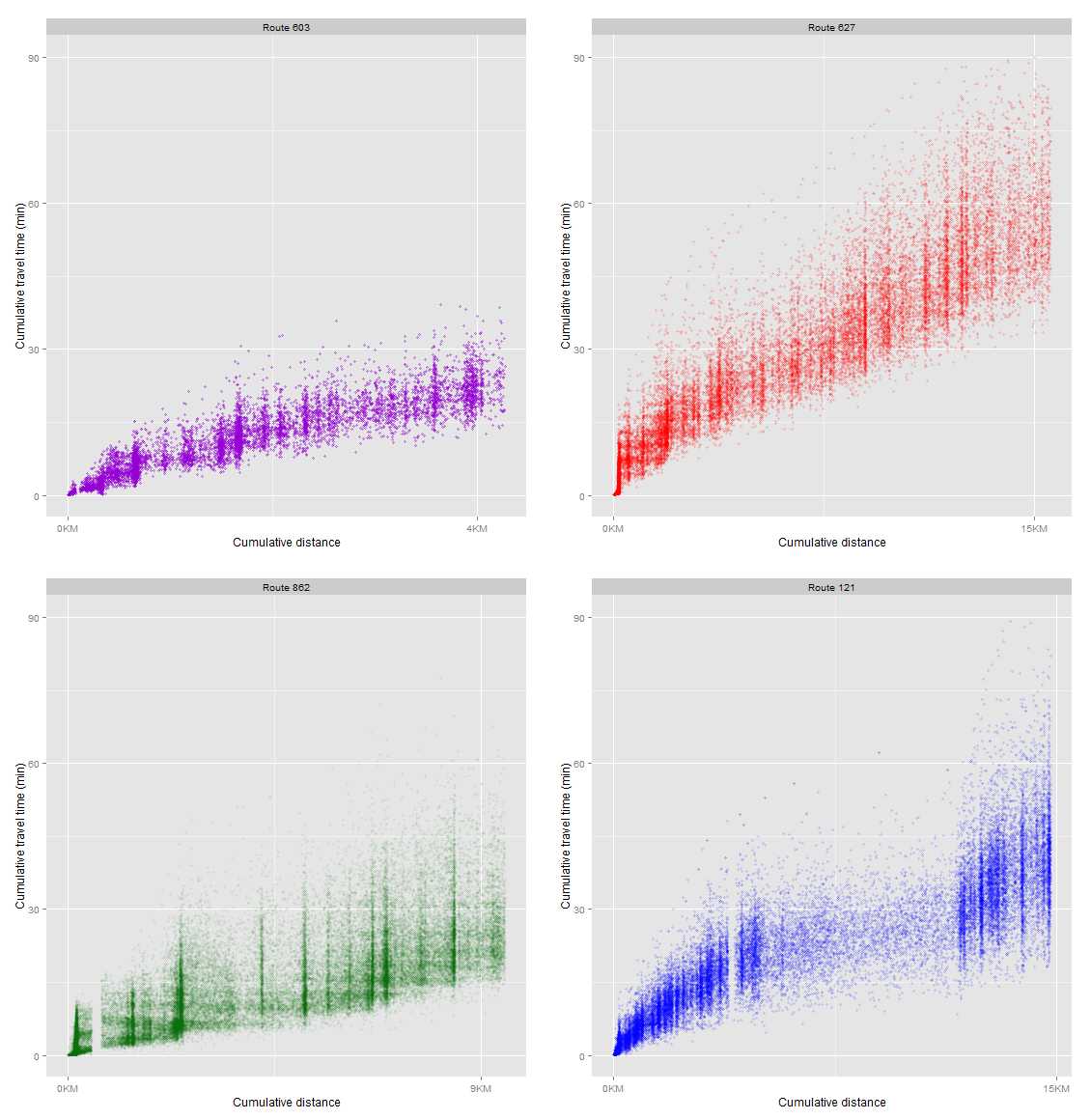}
	\caption{Cumulative space-time trajectories of the four bus routes analyzed in this paper. For each route a different transparency factor between $0$ (completely transparent) and 1 (completely opaque) was chosen for plotting. This was done to normalize for varying numbers of data points per route and further make it possible to see where the bulk of the points lie. Note the distinct x-scales that reflect the different route lengths.}
	\label{fig:rawgpsdata}
\end{figure}

We also had access to GTFS (General Transit Feed Specification\footnote{http://developers.google.com/transit/gtfs}) data, which contain general information about the bus routes, such as bus stop locations. In general, each route consists of two trips, one going from origin to destination and the second representing the return trip. The GTFS data contain a complete definition of each such trip as a sequence of latitude/longitude points tracing the streets of the route from origin to destination. In Figure~\ref{fig:Routeggmap_all} we display on map the bus stops and piecewise linear representations of routes 121 (running from Copacabana to City Center), 603 ( running from Saens Pena to Usina), 627 (running from Saens Pena to Inhauma) and 862 (running from Jacarepagua to Barra da Tijuca).

Note that the observed data did not present itself without any challenges. For example, for each bus entry we only observe a general route ID for the round trip. The GPS data provide no further information about which direction the buses are travelling. However, by analyzing consecutive GPS measurements it is possible to infer the bus direction on the route. Other challenges involved erroneous or non-informative data entries. For example, for some buses the GPS measurements were observed far from the given routes and even at remote locations. We systematically removed all such non-informative entries in subsequent analyses.

\subsection{Data Normalization}

The GPS data in conjunction with the GTFS data provide us with the means to map GPS coordinates onto a 1-dimensional scale measuring distance from origin. For any given bus coordinate we project it onto the closest line segment of the corresponding route and then calculate its distance from origin along the piecewise segments. 

By calculating differences between consecutive time stamps we may infer travel times of each bus between its observed locations. However, in order to analyze and compare travel times of buses, running at different hours, we need to normalize the time stamps onto a common cumulative time scale, i.e., we need to define a common time zero. This may be achieved by interpolating all the observed space-time trajectories at a common fixed point in space, e.g., origin, and defining that point as time zero. Space-time trajectories whose GPS coordinates have been mapped onto a cumulative distance scale and whose time stamps have been normalized to a common cumulative time scale are called \emph{cumulative space-time trajectories}.

In Figure~\ref{fig:rawgpsdata}, we see the cumulative space-time trajectories of all buses (during the specified time period) running on the four routes analyzed in this paper. Note that the only interpolation made is at origin to define the common cumulative time scale. In all other aspects, the scatter plots represent raw measurements observed at irregular spatial locations. 

\subsection{Mathematical Notation} \label{subsec:cumulativespacetimetraj}

In general, we may normalize the time stamps at any arbitrary fixed point in space, in particular, at any of the route's bus stops. Let $0=p_0 < p_1 < \dots < p_K$ denote the distances of all bus stops of a given route from origin $p_0$, where $K$ denotes the number of on-route bus stops. \emph{Cumulative space-time trajectories normalized at $p_k$} consist of cumulative distances $p_k \leq \textrm{\emph{dist}}_{ijk} \leq p_K$, and corresponding cumulative travel times $T_{ijk} \geq 0$, $j=1,\dots,m_{ik}$, where $m_{ik}$ denotes the number of data points for bus $i$ beyond $p_k$. The distances may either represent interpolated values at prespecified fixed locations (e.g., subsequent bus stops) such as in~\cite{lee2012http,sinn2012predicting}, or raw GPS coordinates as in this paper.
Both cumulative distances and cumulative times are defined from $p_k$ onward such that $\textrm{\emph{dist}}=0$ and $T=0$ at $p_k$. The cumulative time scale is inferred by interpolating two consecutive time stamps before and after $p_k$. We denote by \emph{Traj}$(p_k)$ the set of thus normalized historical cumulative space-time trajectories.

\section{Theoretical Background} \label{sec:background}

Additive models~\cite{Hastie,Wood} are linear models, which allow the linear predictor to not only depend on pure linear terms but also on a sum of unknown smooth functions of predictor variables. This class of models is particularly powerful when there is an evident smooth relationship between the response and predictor variables but exact parametric form can neither be theoretically nor intuitively inferred. However, we need to specify these functions in some meaningful way and determine the degree of smoothness. This section discusses both of these topics.

\subsection{Penalized Spline Smoothing}

Let us first consider one-dimensional functions through the simple scatterplot smoothing model 
\begin{equation}
y_i = f(x_i)+\varepsilon_i, 
\end{equation} 

\noindent $i=1,\dots,n$. A common approach~\cite{Wood,Ruppert}, is to represent the function as $f(x)=\sum_{j=1}^q \beta_j \phi_j(x)$, where $\phi_j(x)$ are known basis functions and $\beta_j$ are coefficients to be estimated. An intuitive example is the piecewise linear representation, involving basis functions $\phi_1(x)=1$, $\phi_2(x)=x$, and $\phi_{j+2}(x)=(x-\tau_j)_+ \equiv\max (0,x-\tau_j)$, $j=1,\dots,q-2$, where $\tau_j$ are called knots that need to be chosen (e.g., equally spaced in $x$-domain). The exact choice of the number of knots and placement is not generally critical and is not the focus of this paper. In general the number should be chosen to be large enough to represent the underlying truth reasonably well, while at the same time maintaining computational efficiency. By letting $X=[1 \textrm{ } x_i \textrm{ } (x_i-\tau_1)_+ \textrm{ }\dots \textrm{ } (x_i-\tau_{q-2})_+]_{1 \leq i \leq n}$ the model function may now be written in matrix form as $f(x) = X \beta$. This representation is quite general and there exist several families of basis functions that fit into the above framework. For example, a simple cubic regression spline can be obtained by defining $\phi_{j+2}(x)=|x-\tau_j|^3$ instead of the truncated linear basis above.

The above model may be estimated by least squares or by maximizing the loglikelihood function, 
$\ell(\beta)$, under a normality assumption on $\varepsilon$. However, in order to control the smoothness of the fit we need to work with the so called penalized loglikelihood
\begin{equation}
\ell_P = \ell(\beta) - \lambda \beta' D \beta, \label{eqn:penalizedloglikelihood}
\end{equation}

\noindent where $D$ is most often specified as $\textrm{diag}(0,0,1,\dots,1)$ and $\lambda$ is a smoothness parameter. If $\lambda$ is chosen too large the resulting fit becomes closer and closer to a linear fit in the above case. On the other hand, choosing $\lambda$ too small may lead to overfitting. In general the smoothness parameter may be estimated, for example, using Generalized Cross Validation (GCV).

\vspace{5pt}
\subsection{Additive Models}

The additive models that we consider in this paper have the form:
\begin{equation}
y_i = X_{0i}\beta_0 + f_1(x_{1i})+f_2(x_{2i})+f_3(x_{1i},x_{2i}) + \varepsilon_i, \label{additive.model}
\end{equation}
where $y_i$ and $\varepsilon_i$ are the response and error term respectively, $X_{0i}\beta_0$ represents purely linear terms in the model, and $f_1$, $f_2$, and $f_3$ represent smooth functions of the predictors $x_1$ and $x_2$. We represent the one-dimensional functions as in the previous subsection: $f_1(x_1) = \sum_{j=1}^{q_1} \beta_{1j} \phi_j(x_1)$ and $f_2(x_2) = \sum_{k=1}^{q_2} \beta_{2k} \psi_k(x_2)$, where $\phi_j(x_1)$ and $\psi_k(x_2)$ are known (possibly distinct) basis functions. In this paper a tensor product basis~\cite{Wood,Ruppert} is used to represent the bivariate term:
\begin{equation}
f_3(x_1,x_2) = \sum_{j=1}^{q_1} \sum_{k=1}^{q_2} \beta_{3jk} \phi_j(x_1)\psi_k(x_2).
\end{equation}
Through a similar argument as in the previous subsection, each of the above functions $f_i$, $i=1,2,3$, may be represented by $X_i \beta_i$, where the $X_i$ matrices are appropriately specified in terms of the basis functions $\phi_j(\cdot)$, and $\psi_k(\cdot)$. The model terms may then be stacked in the traditional way: $X=[X_0 \textrm{ } X_1 \textrm{ } X_2 \textrm{ } X_3]$, and $\beta=(\beta_0',\beta_1',\beta_2',\beta_3')'$, to obtain the linear model:
\begin{equation}
Y = X \beta + \varepsilon. \label{eqn:mixedmodelrepresentation}
\end{equation}
This demonstrates that with the appropriate specification of the smooth functions an additive model is simply a linear model whose smoothness of fit may be controlled by placing a penalty on the $\beta$ terms. We may separately control the smoothness of each function by introducing function specific smoothness parameters $\lambda_i$. The penalized loglikelihood from (\ref{eqn:penalizedloglikelihood}) then generalizes naturally to:
\begin{equation}
\ell_P = \ell(\beta) - \sum_{i=1}^3 \lambda_i \beta_i' D_i \beta_i, \label{eqn:penalizedlikelihood}
\end{equation}
where $D_i$ are specified similarly.

Estimation of the additive model may be performed by maximizing the penalized likelihood in (\ref{eqn:penalizedlikelihood}) and estimating the smoothness parameters through GCV. Once the model has been estimated using training data, one can predict a new response in the usual manner.

\subsection{Additive Model with Random Intercept}\label{sec:random_intercept}

In this paper we also consider an additive model with a random intercept
 \begin{equation}
y_i = b_{0i} + X_{0i}\beta_0 + f_1(x_{1i})+f_2(x_{2i})+f_3(x_{1i},x_{2i}) + \varepsilon_i, \label{additive.mixedmodel}
\end{equation}
where $b_{0i} \sim N(0,\sigma_b^2)$ and $\varepsilon_i \sim N(0,\sigma_\varepsilon^2)$. Note that the above model is not overparametrized as the $b_{0i}$ are treated as random and not fixed. This model falls into the general class of Additive Mixed Models~\cite{Wood} (due to the mixed combination of random and fixed model terms) and we note that by specifying the smooth functions as before it may be represented in the matrix form:
\begin{equation}
Y = X \beta + Z b_0 +\varepsilon, \label{eqn:mixedmodel}
\end{equation}
 where $Z$ is a single column matrix of ones. 

The estimation of the above model is not straight forward and since space is limited we point to~\cite{Wood} for full theoretical coverage. However, we note that through a correct likelihood specification an iterative maximization algorithm may be applied to obtain estimates of the parameters $\beta$, $\sigma_b^2$, and $\sigma_\varepsilon^2$. Given these estimates the prediction formula for the random effect vector is (see e.g.~\cite{Ruppert})
\begin{equation}
\hat{b}_0 = \sigma_u^2 Z'V^{-1}(y-X\beta), \label{eqn:BLUP}
\end{equation}
where $V = \sigma_u^2 Z Z' + \sigma_\varepsilon^2 I$.

\subsection{Computational Aspects}

Additive models, such as (\ref{additive.model}) using penalized splines and tensor product smooths are implemented in a highly optimized R-package, \emph{mgcv} which allows estimation of the model, \cite{Wood1,Wood2,Wood3,Wood4}. Additive mixed models, such as (\ref{eqn:mixedmodel}) are more computationally expensive than regular Additive Models, in particular when the number of random effects becomes large. However, the \emph{mgcv}-package also has a optimized routine for estimation through a call to the \emph{lme} function of the highly developed \emph{nlme} R-package \cite{pinheiro} that was specifically designed to estimate linear mixed models efficiently.

\section{Proposed Solution}\label{sec:modeling}

\begin{figure}[!t]
	\centering
	\includegraphics[width=\linewidth]{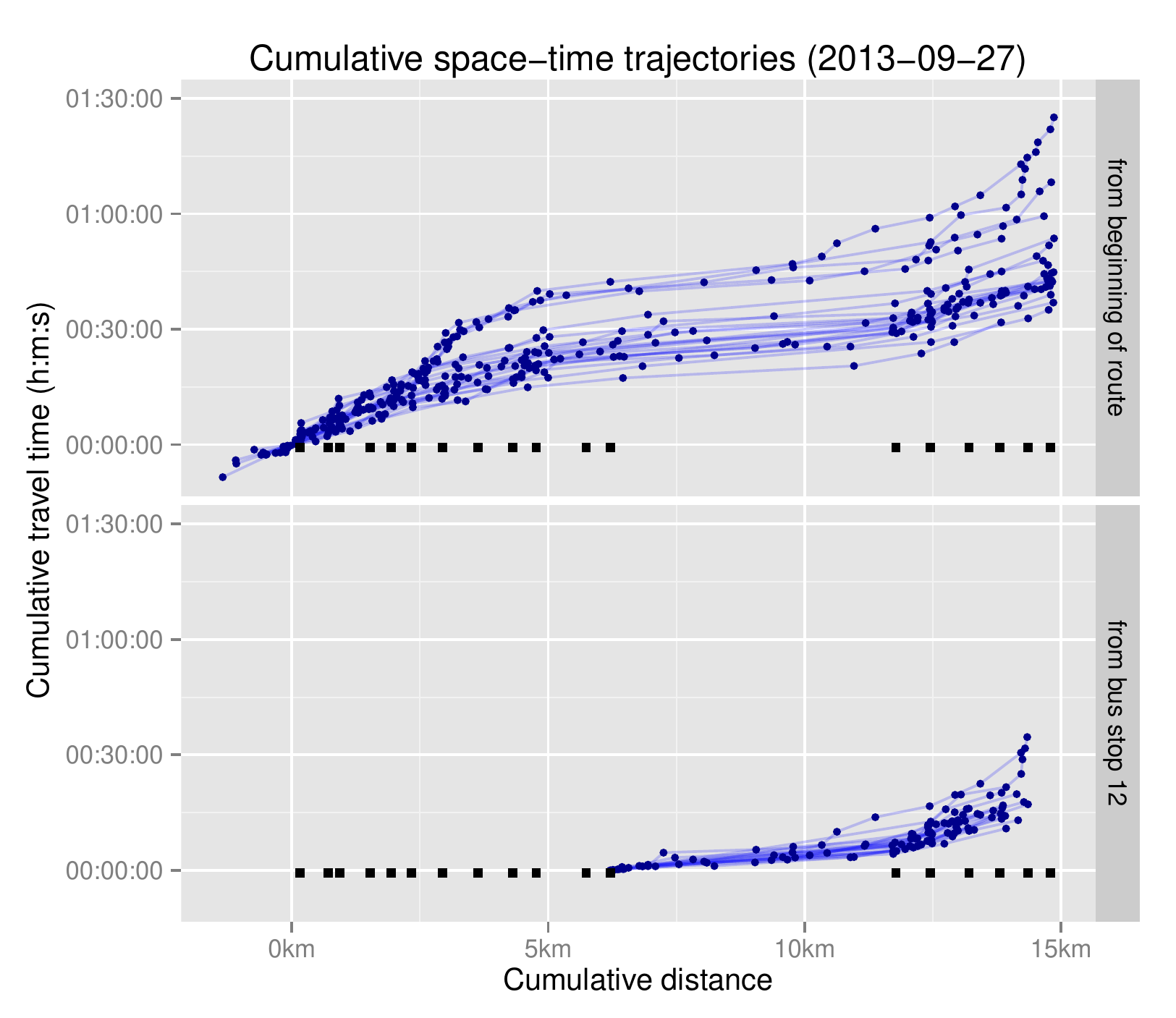}
	\caption{Cumulative space-time trajectories for route 121 from September 27, 2013. Upper panel demonstrates the trajectories from beginning of route and the lower panel from bus stop $p_{12}$ onward. Bus stops are marked by black squares.}
	\label{fig:SpaceTimeTraj}
\end{figure}

In this section we present additive models for analyzing historical cumulative space-time trajectories such as those observed in Figure~\ref{fig:SpaceTimeTraj}. In the upper and lower panel, respectively, we see examples of historical trajectories, \emph{Traj}$(p_k)$, that have been normalized at $p_k$ for $k=0$ (origin) and $k=12$ (bus stop $p_{12}$). We note that the cumulative travel time variance beyond bus stop $p_{12}$ is reduced dramatically when normalized at $p_{12}$ as compared to at $p_0$. Therefore, we propose to train additive models on each of the historical trajectories, \emph{Traj}$(p_k)$, for bus stops $k=0,\dots,K-1$, where $K-1$ corresponds to the second to last bus stop on route. The objective is then to base future travel time predictions of a bus close to bus stop $p_k$ on the corresponding additive model trained on \emph{Traj}$(p_k)$. 

In order to make our presentation more coherent, we model and analyze the bus trajectories of bus route 121 (Copacabana-Center) for the first two weeks of our observed time period. We analyze trajectories starting from origin, \emph{Traj}$(p_0)$ and for ease of notation we omit the $k$-subindex of Subsection~\ref{subsec:cumulativespacetimetraj}. However, we note that all discussions generalize to trajectories starting from any given bus stop along the route, \emph{Traj}$(p_k)$. For the first two weeks of our study we observed $n=385$ trajectories for route 121 with on average $m_i=13$ measurements per bus ride. Through statistical reasoning, we construct three models whose performances are compared to previous approaches (Section~\ref{sec:experiments}). All numerical summaries in this section apply to this data set.

\begin{figure}[!t]
	\centering
	\includegraphics[width=\linewidth]{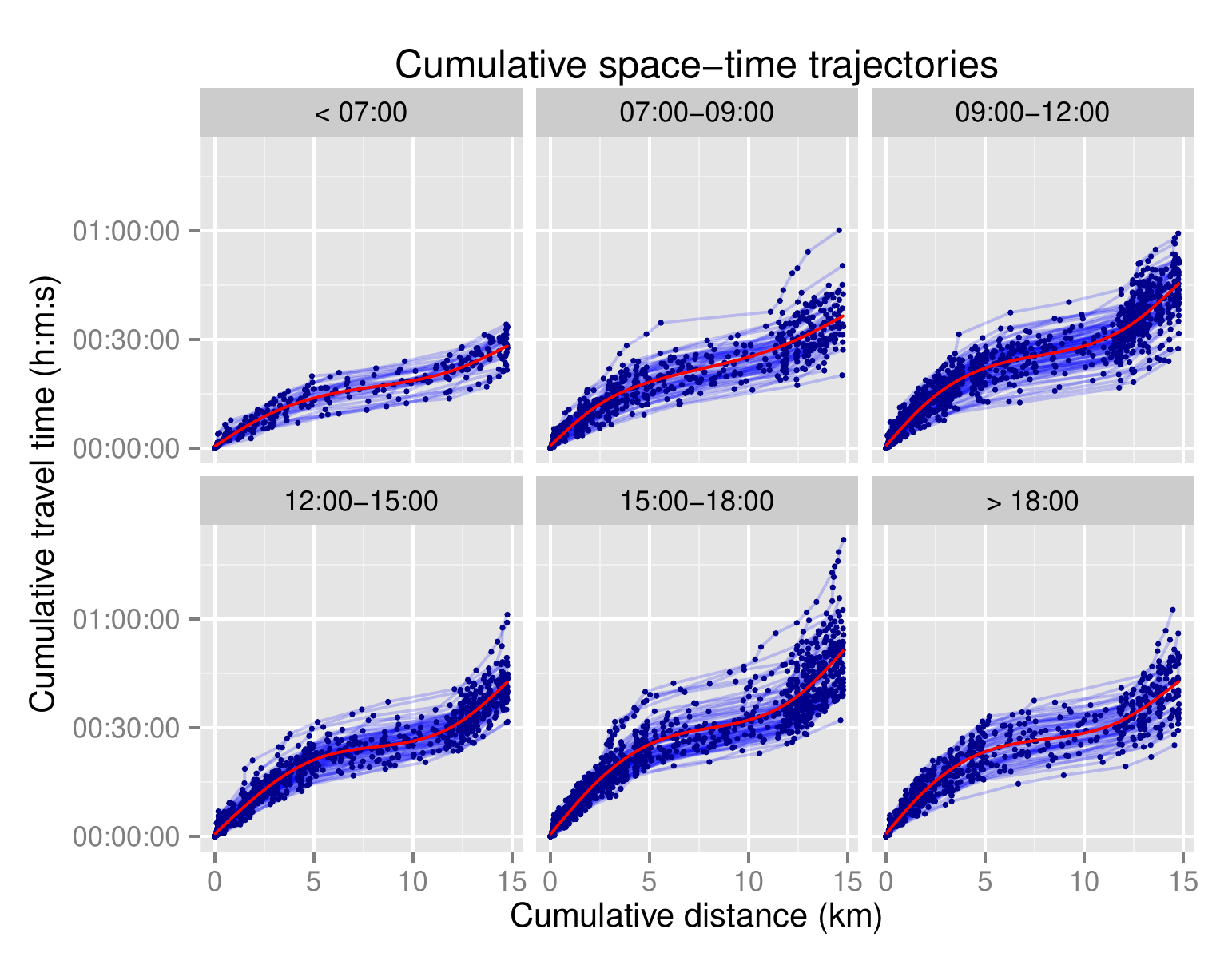}
	\caption{Cumulative space-time trajectories of route 121, stratified by hour, during the time period September 26 - October 10, 2013. The dots represent raw measurements and the blue interpolated curves represent each bus trajectory for illustration purposes. A smooth mean curve for each time category is depicted in red.}
	\label{fig:stratification}
\end{figure}

\subsection{Basic Additive Model for Travel Times}

In Figure~\ref{fig:stratification}, we see the cumulative space-time trajectories of all bus rides of route 121 during the specified time period. The trajectories are stratified by hour and a smooth mean curve is fitted through each scatterplot to illustrate travel time trends. We note that morning travel time duration peaks between 9am and noon (morning rush hour). Then a slight reduction in travel times is observed between noon and 3pm, followed by an afternoon rush hour. We also note that there is not only a difference in total travel times across hours, but also in the shapes of the mean curves. This figure inspires the following model of bus travel time, $T_{ij}$, as function of distance from origin, \emph{dist}$_{ij}$, and time of departure, \emph{time}$_{i}$:
\begin{align}
& \textrm{\textbf{Model 1: Basic Additive Model (BAM)}} \nonumber \\
& T_{ij} = \beta_0 + f_1(\textrm{\emph{dist}}_{ij})+f_2(\textrm{\emph{time}}_{i})+f_3(\textrm{\emph{dist}}_{ij},\textrm{\emph{time}}_{i}) +  \varepsilon_{ij}, \nonumber
\end{align}
$i=1,\dots,n$, and $j=1,\dots,m_i$, where $\beta_0$, and $\varepsilon_{ij}$ represent an overall model mean and error term, respectively. In what follows we assume the random error terms are mean zero and normally distributed. The terms $f_1$, $f_2$, $f_3$ denote unknown smooth functions designed to capture functional relationships such as those observed in Figure~\ref{fig:stratification}. The $f_1,f_2$-terms can be thought of as smooth main effects of \emph{dist} and \emph{time} on $T$, respectively, whereas the $f_3$-term represents an interaction effect of the two variables. The interaction allows the functional relationship between $T$ and \emph{dist} to change with \emph{time}, as observed in Figure \ref{fig:stratification}. 

The functions $f_1$, $f_2$ and $f_3$ were represented by cubic regression splines and tensor product smooths (see Section~\ref{sec:background}). We placed one knot at each bus stop between origin and destination to capture smooth transitions from one station to the next. We placed $5$ equally spaced knots in the time space, which was large enough to capture the two rush hours trends in the morning and afternoon, respectively. Larger number of time-knots did not seem to affect the fit of the model.

\begin{figure}[!t]
	\centering
	\includegraphics[width=\linewidth]{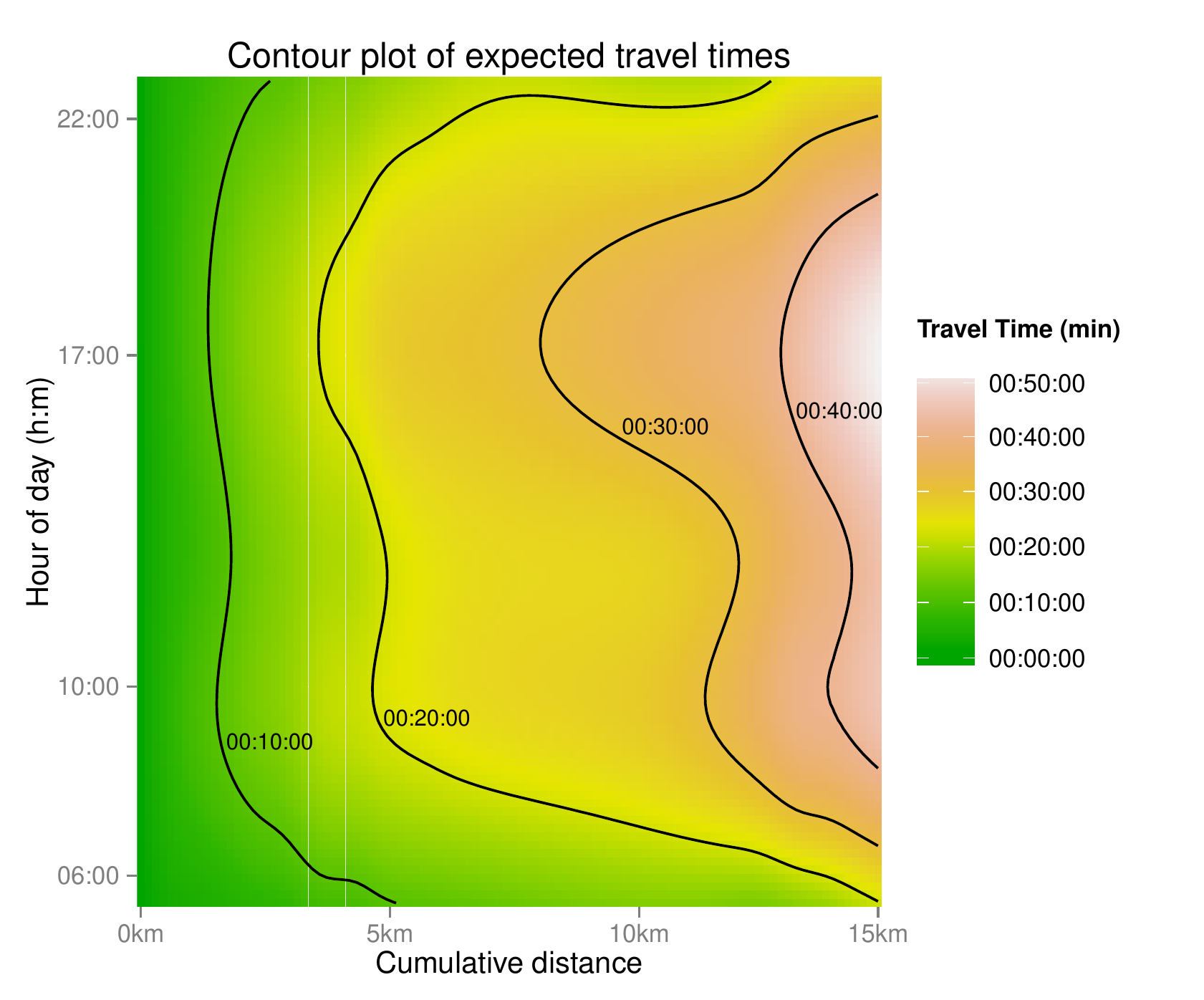}
	\caption{A contour plot of estimated travel times as a smooth function of time of day and cumulative distance from origin.}
	\label{fig:effectplot}
\end{figure}

We estimated the Basic Additive Model using the \emph{mgcv} R package. The numbers showed that each of the functional effects $f_1$, $f_2$, and $f_3$ was deemed statistically significant by the F-test (p-values $< 10^{-16}$) and the overall adjusted $R^2$ of the model was $0.903$. To illustrate the smooth relationship between the two variables and travel time, Figure~\ref{fig:effectplot} shows a contour plot of estimated travel time with cumulative distance from origin on x-axis and time of day on y-axis. We can see that at 10am it takes the bus on average approximately $30$min to travel around $12$km, while at 5pm it only travels around $8$km in half an hour. We also see that the two rush hour peaks, at approximately 10am and 5pm, are more amplified at $12$km than at $2$km, as exemplified by the $30$min and $10$min contour lines, respectively. These observations demonstrate the importance of including the interaction term $f_3$ in the model.

\subsection{Extended Additive Model with Additional Features}

\begin{figure}[!t]
	\centering
	\includegraphics[width=\linewidth]{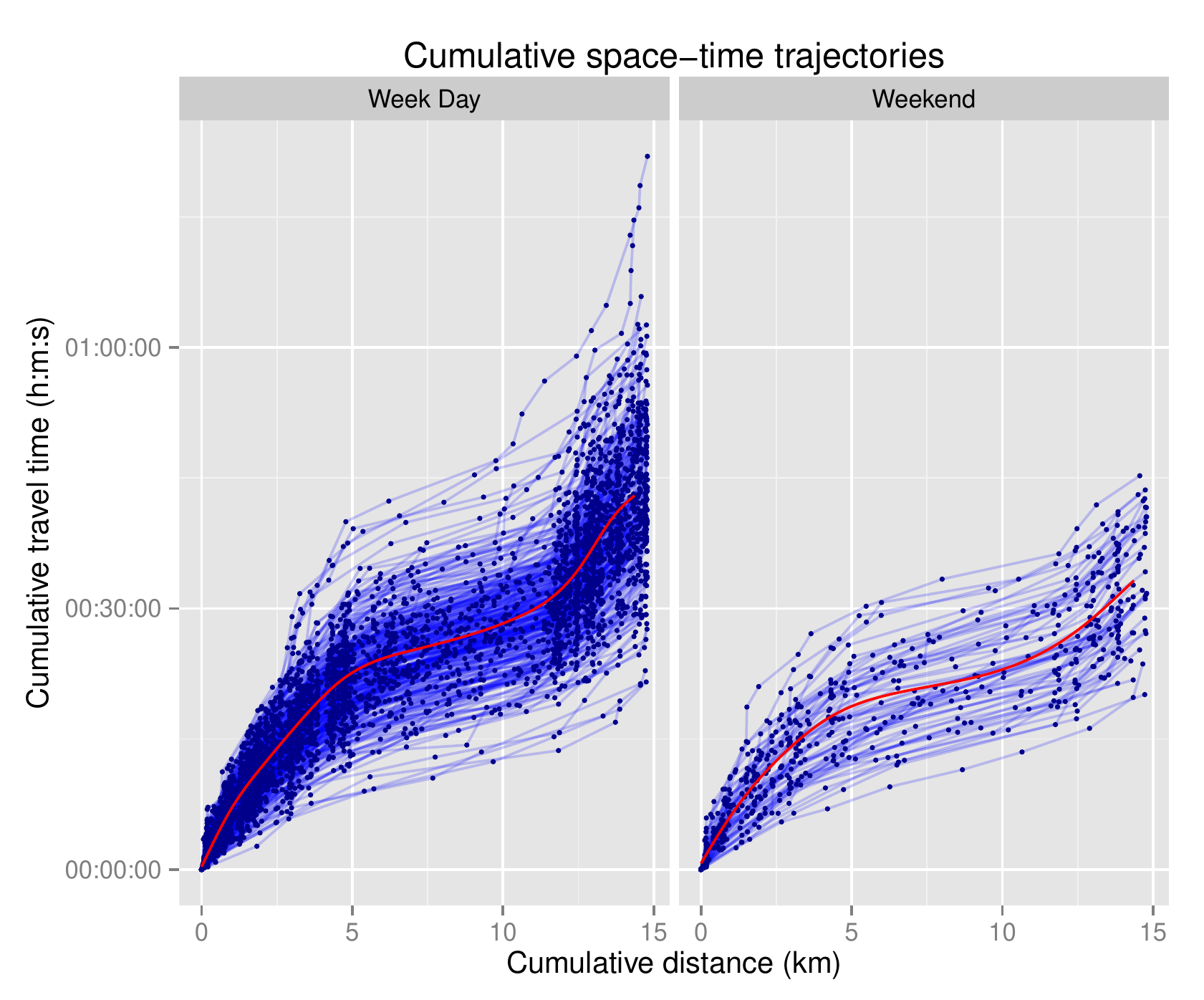}
	\caption{Cumulative space-time trajectories of route 121 during week days (left) and weekends (right). The dots represent raw measurements and the blue interpolated curves represent each bus trajectory for illustration purposes. Smooth mean curves are depicted in red.}
	\label{fig:weekendeffect}
\end{figure}

It is well known that traffic patterns in cities are different on a weekday as compared to the weekend (see Figure~\ref{fig:weekendeffect}). This phenomenon can easily and quite flexibly be incorporated into our basic additive model above. Let \emph{weekend}$_i$ denote an indicator variable that determines whether bus ride $i$ occurred on a weekday or on the weekend. 
Then by adding the linear term $\beta_1 \cdot \textrm{\emph{weekend}}_i$ into the model we account for differences in overall mean travel times between weekdays and weekends. However, there is an evident interaction of the weekend factor with distance from origin as can be seen in Figure~\ref{fig:weekendeffect}, where the mean difference between weekday and weekend travel times 
increases as a function of distance. Therefore, in addition to the main effect $\beta_1 \cdot \textrm{\emph{weekend}}_i$, we propose replacing the functional term $f_1(\textrm{\emph{dist}}_{ij})$ in Model 1 by the interaction term $f_1(\textrm{\emph{dist}}_{ij},\textrm{\emph{weekend}}_i)$. This term in fact generates two different smooths, one for weekday and the other for weekend 
trajectories.

Another feature that intuitively seems likely to correlate well with travel time of a given bus is the travel time of the last bus in front of it. We therefore define the feature 
$T_{ij}^{\textrm{last}}$ to be the cumulative travel time at \emph{dist}$_{ij}$ of the last bus that passed before time of departure of bus $i$. It is important to point out here 
that it is unlikely that the last bus will transmit a GPS signal at the same locations \emph{dist}$_{ij}$ as bus $i$. Therefore, interpolation of the cumulative space-time trajectory of the last bus 
is performed at \emph{dist}$_{ij}$ to construct the feature $T_{ij}^{\textrm{last}}$. We observed that $T_{ij}^{\textrm{last}}$ had a strong linear relationship with the observed travel times, $T_{ij}$, with a sample correlation of $r\approx0.92$.

The following model extends Model 1 to include the features discussed above:
\begin{align}
& \textrm{\textbf{Model 2: Extended Additive Model (EAM)}} \nonumber \\
& T_{ij} =  \beta_0 + \beta_1 \cdot \textrm{\emph{weekend}}_i  + f_1(\textrm{\emph{dist}}_{ij},\textrm{\emph{weekend}}_i) \nonumber \\ & \left. \right. \quad +  \beta_2\cdot T_{ij}^{\textrm{last}} + f_2(\textrm{\emph{time}}_{i}) + f_3(\textrm{\emph{dist}}_{ij},\textrm{\emph{time}}_{i}) +  \varepsilon_{ij}. \nonumber
\end{align}

We fitted the above model to our data set and observed that all effects, including \emph{weekend}$_i$, the interaction term $f_1(\textrm{\emph{dist}}_{ij},\textrm{\emph{weekend}}_i)$, and the linear predictor $T_{ij}^{\textrm{last}}$ were highly significant (p-values $< 10^{-16}$). The adjusted $R^2$ increased from the previous model to $0.919$. 

\subsection{Additive Mixed Model}

We recall that in Section~\ref{subsec:cumulativespacetimetraj} we normalized all the space-time trajectories to a common cumulative time scale. Since the actual times of departure are not known, an approximation was made by taking two consecutive time stamps before and after origin and defining $T= 0$ as the interpolated time-stamp at origin. However, note that this introduces an error in the form of a vertical trajectory shift, due to incorrect specification of time of departure. One bus driver may, for example, take a short break at origin, while another may depart soon after arriving. This error is unpredictable in our context and can amplify when time resolution is poor such as in our data, where GPS coordinates are only transmitted on average every 4 minutes. 

In order to correct for the misspecification of time of departure, we propose an additive mixed model (see section~\ref{sec:random_intercept}) that includes a (corrective) random intercept term $b_{0i}$ for each and every bus ride $i=1,\dots,n$:
\begin{align}
& \textrm{\textbf{Model 3: Additive Mixed Model (AMM)}} \nonumber \\
& T_{ij} =  \beta_0 + b_{0i} + \beta_1 \cdot \textrm{\emph{weekend}}_i  + f_1(\textrm{\emph{dist}}_{ij},\textrm{\emph{weekend}}_i) \nonumber \\ & \left. \right. \quad +  \beta_2\cdot T_{ij}^{\textrm{last}} + f_2(\textrm{\emph{time}}_{i}) + f_3(\textrm{\emph{dist}}_{ij},\textrm{\emph{time}}_{i}) +  \varepsilon_{ij}, \nonumber
\end{align}
where $b_{0i} \sim N(0,\sigma_b^2)$. 

We fitted the above model and found that the random intercept term $b_{0i}$ was indeed highly significant (p-value $< 10^{-16}$) and the adjusted $R^2$ value increased 
significantly to $0.968$. The standard deviation $\sigma_b$ was estimated to be $3$ minutes, which indicates that the estimated (interpolated) time of departure indeed requires adjustment.

\section{Experiments}\label{sec:experiments}

\vspace{5pt}
\subsection{Experimental Data}

We performed a prediction analysis on four bus routes in the city of Rio de Janeiro: 603, 627, 862, and 121 (see Figure~\ref{fig:Routeggmap_all}). These routes are located in distinct regions of the city and further have different lengths, number of bus stops, and frequency of bus rides, see Table \ref{table_routes}. Further these routes demonstrate distinct traffic patterns as can be seen in Figure~\ref{fig:rawgpsdata}.
%To ensure that our experiments are easily reproducible, we have built a website which contains all the data sets 
We have made all the data sets available in~\cite{dataset}, which we believe this will stimulate further research in the area.

We note that since GPS locations are transmitted on average every 4 minutes the density of points in the spatial dimension of Figure~\ref{fig:rawgpsdata} provides some insight into traffic behavior at different route segments. We see, for example, a lighter blue section in the middle of route 121 (due to fewer observations), which represents an inner city thruway less prone to traffic congestions. On the other hand, darker sections represent locations where traffic may experience regular stops or delays, such as traffic signals or frequently congested road segments. Locations with no data points represent either tunnels or regions with poor reception. We note that although both routes 121 and 627 have the same length their cumulative space-time trajectories are quite different. 
%In particular those of route 627 look more linear relationship. 
These four distinct routes represent a wide range of prediction scenarios we want to cover in our experiments.

\begin{table}[!t]
\renewcommand{\arraystretch}{1.3}
\caption{Route data summary}
\label{table_routes}
\centering
\begin{tabular}{|c|c|c|c|} \hline
Route & \# trajectories & \# stops & Length (in km) \\
\hline
 603 & 1,276 & 15 & 4 \\\hline 
 627 & 1,325 & 54 & 15 \\\hline
 862 & 7,882 & 24 & 10 \\\hline
 121 & 2,515 & 18 & 15 \\\hline
\end{tabular}
\end{table}

\subsection{Experimental Setup}

The total number of trajectories in each route is presented in Table~\ref{table_routes}.
We randomly selected 14 days after November 1st 2013 as our test data. This guaranteed at least 30 days of historical data for each test date. For each bus running on any of these 14 days we performed travel time predictions using three sets of historical data involving all bus rides in the last 10, 20, and 30 days, respectively. This was done to get a sense of whether the size of the historical data set has an influence on the accuracy of the tested models.

Travel time predictions were made for each bus in test set from every bus stop until end of route to reflect the real world problem of predicting bus arrivals from any on-route location onward. More precisely, for each bus stop $p_k$ we recorded for bus $i$ in test set the first observed bus entry \emph{dist}$_{i1k}$ after $p_k$. We then made travel time predictions at all remaining (observed) points \emph{dist}$_{ijk}$, $j=2,\dots,m_{ik}$. Since the data at \emph{dist}$_{ijk}$ represent the raw data whose cumulative travel times $T_{ijk}$ (from bus stop $p_k$) are known we could thus calculate and analyze prediction errors. In order to get a sense of how error changes as a function of distance from the bus stop beyond which predictions were made we recorded the prediction distances $|\textrm{\emph{dist}}_{ijk}-p_k|$. 

\subsection{Evaluation Measures}
To evaluate overall performance of each method for a given bus route we calculated the \emph{mean absolute relative error}, defined as $(1/N)\sum_{ij} |T_{ij}-\hat{T}_{ij}|/T_{ij}$, where $N$ denotes total number of predictions made. Since the distributions of the relative errors was right skewed in all cases a median could have been used instead of mean. However, as the mean is less robust to outliers it may also provide insight about worst case errors. Overall conclusions were not affected by replacing the mean with median. We performed a non-parametric paired Wilcoxon test to compare the overall performances between methods. 

Since error was greater at later parts of route, we also analyzed the distributions of absolute errors stratified by prediction distances, $|\textrm{\emph{dist}}_{ijk}-p_k|$. The distance space was binned into one kilometer bins [0,1), [1,2), [2,3), ... etc. Visual comparison of distributions was performed using boxplots and a 95th percentile curve; see Figure~\ref{fig:jointboxplotanalysisExtended}.  Since absolute errors were right skewed for each method we performed a two-sided non-parametric paired Wilcoxon test to compare methods within each distance bin. 

To account for multiple testing, p-values were recorded for each comparison and then adjusted using the Benjamini-Hochberg method~\cite{benjaminihochberg}. Statistical significance was determined if adjusted p-values were $< 0.05$.

\subsection{Implemented Methods}

\vspace{2pt}
\textbf{Additive Models:} No model selection or parameter tuning was performed during the training. Instead for each and every training set we estimated the exact same three models as defined in Section~\ref{sec:modeling}. Once estimation had been performed the estimated model parameters, $\hat{\beta}$, along with a complete set of test features was plugged into the Additive Model formula (\ref{eqn:mixedmodelrepresentation}) to obtain travel time predictions at subsequent route locations.
For AMM, in order to estimate the random effect $b_{0i}$ of (\ref{additive.mixedmodel}) for a new trajectory $i$ in the test set at least one observed travel time is needed. Since predictions are always made given the current location of the bus, the first observation, $T_{i1}$, may be used for that purpose. By plugging this value in for $y$ in the formula (\ref{eqn:BLUP}) we obtain an estimate of $b_{0i}$. Then the formula (\ref{eqn:mixedmodel}) may be used in conjunction with a complete set of test features to obtain travel time predictions at subsequent route locations. A minor implementation detail we want to point out involves predictions beyond bus stops very close to route destination. In this case the training data \emph{Traj}$(p_k)$ can become scarce and full spline function representation as defined in our three proposed models in Section~\ref{sec:modeling} may lead to overfitting. Therefore, in these cases, we replaced the smooth functions with the more simple linear model terms: $\alpha_1 \textrm{\emph{dist}}_{ij} + \alpha_2 \textrm{\emph{dist}}_{ij} \cdot \textrm{\emph{weekend}}_{i} + \alpha_3$\emph{time}$_{ij} + \alpha_4 \textrm{\emph{dist}}_{ij} \cdot time_{ij}$.

\textbf{Support Vector Machine (SVM):} Bin et al.~\cite{bin@jits2006} used support vector machine regression to predict the arrival time of the next bus. They divided the bus trajectories in segments and then used as features the travel time of current bus at previous segment and the latest travel time of a previous bus in the next segment to predict the travel time for the next segment. Since we are not only interested in predicting the travel time of the next segment but all subsequent segments until end of route, we added to the training data the latest travel times at all subsequent segments. Similar to~\cite{bin@jits2006}, we used a linear kernel and the implementation was performed using the R package ``e1071''. 

\textbf{Kernel Regression:} Sinn et al.~\cite{sinn2012predicting} proposed an instance-based method that uses weighted averages of historical trajectories to make predictions. Trajectories with similar behaviour up to the current bus location are given more weight. Weights are defined by a gaussian kernel: $\exp(-\|x-y\|^2/b)$, where $x$ and $y$ are cumulative space-time trajectories, and $b$ is the bandwidth of the kernel\footnote{Similar to~\cite{sinn2012predicting}, we set $b=1$ in our experiments}. For further details, we refer the reader to~\cite{sinn2012predicting}.

Both approaches, SVM and Kernel Regression, perform predictions only at predefined route segments. However, since GPS data consist of irregular points in space, both of these methods relied on interpolation at predefined route locations, such as bus stops in~\cite{bin@jits2006}. We therefore performed interpolation at all bus stops of the route, which in fact resulted in a consensus in training data across all methods. To be more precise, for predictions from bus stop $p_k$ onward, all approaches used as training data the historical space-time trajectories \emph{Traj}$(p_k)$. The key difference is that for SVM and Kernel Regression the cumulative distances $\textrm{\emph{dist}}_{ijk}$, underlying \emph{Traj}$(p_k)$, coincide exactly with subsequent bus stops beyond $p_k$, whereas in our approach they correspond directly with the raw GPS measurements. %Since the only way to calculate true error rates is to make predictions at the observed GPS locations, the predictions (made at bus stops) of SVM and Kernel Regression need to be further interpolated at the observed route locations. %This is clearly a limitation of these methods, however we believe this to be the only meaningful error analysis. Further, as these methods rely on interpolation to infer travel times at predefined locations, in the first place we believe that under the assumptions of those models this approach should not 

\subsection{Experimental Results}

In Table~\ref{table_mape} we see the mean absolute relative errors for each method. The first thing to note is that our Additive Models (BAM, EAM, and  AMM) outperformed the Kernel Regression and SVM in all scenarios. SVM's overall performance was notably worse than any of the other methods. The main comparisons of interest are thus between the Kernel Regression approach and each one of our Additive Models. The Wilcoxon paired test revealed statistically significant differences between the Kernel Regression and all our proposed three Additive Models, in all scenarios. Further, the Wilcoxon paired test revealed that in all scenarios the AMM outperformed all other methods. Another observation from Table~\ref{table_mape} is that the size of the training data does not seem to affect performance of any of the $5$ methods.

\begin{table}[!t]
\renewcommand{\arraystretch}{1.3}
\caption{Mean Absolute Relative Error}
\label{table_mape}
\centering
\small
\begin{tabular}{|c|c|c|c|c|c|c|}
\hline
\multicolumn{2}{|c|}{} & \multicolumn{5}{ |c| }{Method} \\\hline
 Route & \# days & BAM & EAM & AMM & Kernel & SVM \\\hline
\multirow{3}{*}{603} & 10 & 19.9\% & 19.7\% & \textbf{18.4\%} & 21.3\% & 64.4\% \\
  & 20 & 20.1\% & 19.8\% & \textbf{18.5\%} & 21.3\% & 64.7\% \\ 
  & 30 & 19.8\% & 19.6\% & \textbf{18.3\%} & 21.3\% & 64.8\% \\ 
\hline 
\multirow{3}{*}{627} & 10 & 16.3\% & 14.7\% & \textbf{13.8\%} & 18.1\% & 28.8\% \\ 
  & 20 & 15.2\% & 14.2\% & \textbf{13.4\%} & 17.3\% & 30.0\% \\ 
  & 30 & 15.1\% & 14.0\% & \textbf{13.2\%} & 17.1\% & 29.4\% \\ 
\hline 
\multirow{3}{*}{862} & 10  & 22.1\% & 19.5\% & \textbf{18.0\%} & 23.8\% & 26.4\% \\ 
  & 20 & 22.5\% & 19.3\% & \textbf{18.0\%} & 23.6\% & 26.8\% \\ 
  & 30 & 22.2\% & 19.3\% & \textbf{17.9\%} & 23.4\% & 25.6\% \\ 
\hline
 \multirow{3}{*}{121} & 10 & 23.1\% & 20.9\% & \textbf{19.2\%} & 23.9\% & 41.5\% \\ 
  & 20 & 22.9\% & 20.7\% & \textbf{19.1\%} & 23.6\% & 41.4\% \\ 
  & 30 & 22.7\% & 20.3\% & \textbf{18.9\%} & 23.4\% & 41.2\% \\ 
   \hline
\end{tabular}
\end{table}

\begin{figure*}[!t]
	\centering
	\includegraphics[width=\linewidth,height=5.6in]{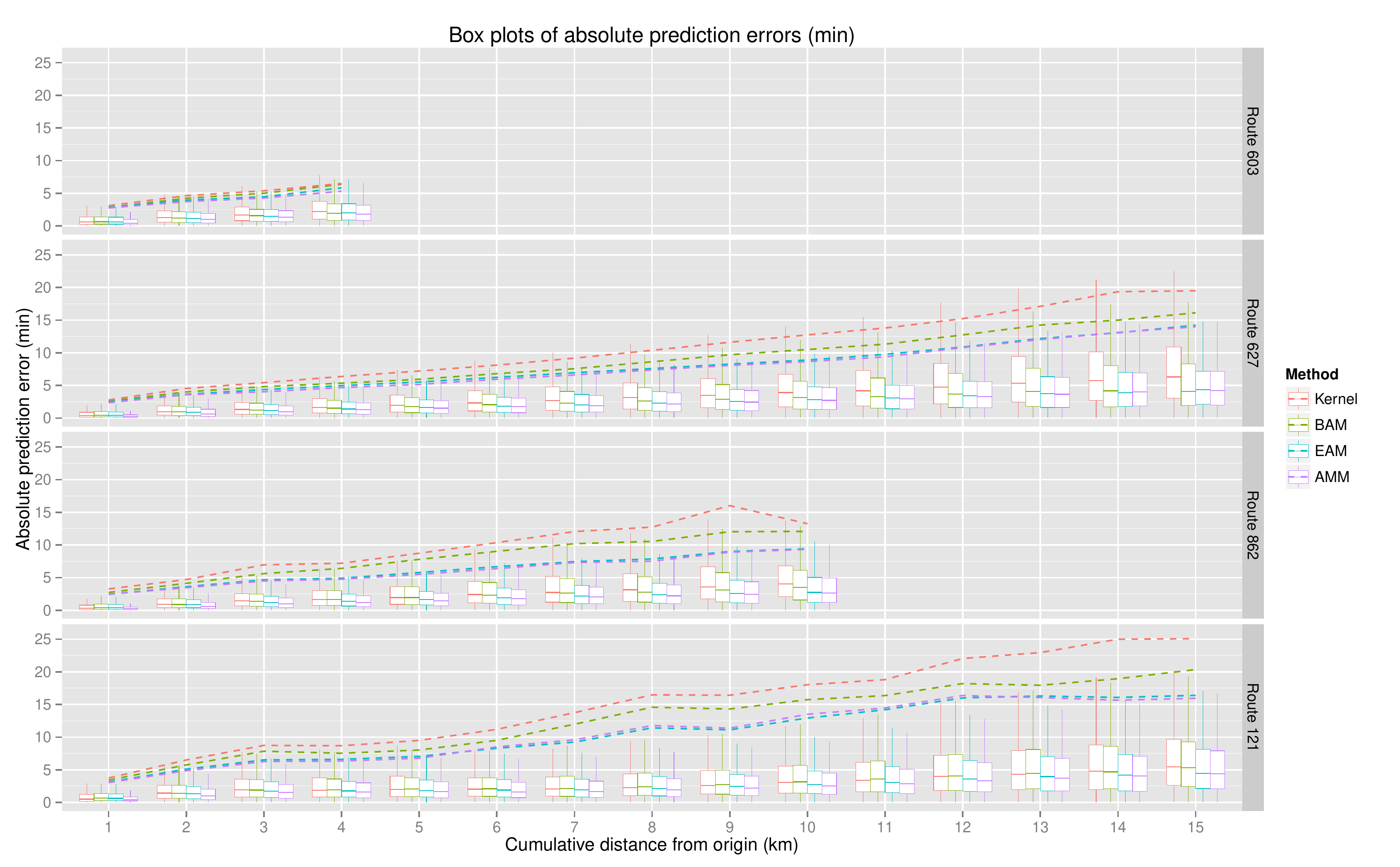}
	\caption{Boxplots of absolute prediction errors for routes 603, 627, 862, and 121, ordered by overall performance of methods. The dashed lines represent 95th percentiles of corresponding absolute errors.}
	\label{fig:jointboxplotanalysisExtended}
\end{figure*}

To give a more detailed view of the results, we show in Figure~\ref{fig:jointboxplotanalysisExtended} boxplots of absolute prediction errors aggregated across all training data sets, $10$, $20$, and $30$ days, and further stratified by route and prediction distance. The boxplots are displayed for all methods except for SVM as their performance was greatly inferior for larger distances and only interfered with visualization. It should be noted that the SVM approach of~\cite{bin@jits2006} was specifically designed to predict only the travel time at next route segment and therefore inferior performance at larger distances may be expected. However, even at the smaller bins the SVM approach was outperformed by all other methods. 

As expected, the error increases with distance from bus stops beyond which the predictions were made. We note that all distributions were right skewed with several outliers (as defined by the ends of the boxplot whiskers) mostly due to heavily delayed buses. These outliers are not displayed as they interfere with visualization and do not reveal any significant trends beyond those seen in the boxplots. However, in order to get a sense of this ``outlier effect'' we plotted the 95th percentiles (dashed lines) along with the boxplots. These lines give us a sense of ``worst case'' scenario performance of each method.

Although perhaps not visually striking in all distance bins, the AMM statistically outperformed all methods for all routes and in all distance bins (except for the 14km bin on route 627 where no difference existed between EAM and AMM). In the first distance bin, $[0,1)$, the Kernel Regression method outperformed both BAM and EAM at all routes except for route 603 (where no statistical difference existed). However, in all other distance bins the two Additive Models statistically outperformed the Kernel Regression. Thus, on the whole, the visualization and stratified analysis confirmed the performance order observed in Table~\ref{table_mape}. 

The fact that Kernel Regression outperformed BAM, and EAM in the first distance bin suggests that the Additive Models tend to put more priority on minimizing error in later parts, when it is in fact larger, at the expense of short term predictions. Perhaps this may be fixed by placing more knots at the beginning of the route or through additional features. However, as we discussed before, AMM performed statistically better than all other methods in the first distance bin. This suggests that the random corrective intercept term plays an important roll in rescuing the incorrectly specified cumulative time scales as obtained by interpolation. 

It is interesting to note that route 121 showed the highest worst-case scenario across all routes, as demonstrated by the 95th percentile curves. This is further reflected in the highest relative error in Table~\ref{table_mape}. This fact can perhaps be explained by the fact that the destination of route 121 lies in the heart of the city center.

\section{Related Work}\label{sec:related}

A list of related works on bus arrival time prediction may be found in~\cite{yu@multipleroutes,Rajbhandari2005,altinkaya}. In this section we present an overview of the main methods, but refer to~\cite{yu@multipleroutes,Rajbhandari2005,altinkaya} for a more exhaustive list of references. The discussion is divided into categories based on the type of models in question.

\subsection{Historical Data-Based Models}

The models falling into this category base predictions of future travel times on historical averages~\cite{sun2007predicting,tiesyte2008similarity,lee2012http,sinn2012predicting}. The proposed algorithm in~\cite{sun2007predicting} combined real-time GPS coordinates and current bus speed with historical average speeds of individual route segments. The methods proposed in~\cite{tiesyte2008similarity,lee2012http,sinn2012predicting} were all based on averages of similar past bus trajectories. These methods work best when current bus has traveled some distance and its trajectory until current location has sufficient data points that can be compared to historical trajectories. In~\cite{sun2007predicting,lee2012http} the analyses were stratified by hour of day by defining time bins. Our proposed BAM falls into this class of models but in addition models temporal effects as a smooth function as opposed to using categorical binning.

\subsection{Regression Models}

Regression models predict and explain a response variable through a function of predictor variables. \cite{jeong2004} and \cite{ramakrishna2006} developed multiple linear regression models using different sets of predictors and both studies indicated that regression models are outperformed by other models. Further, the Kernel Regression method (analyzed in the Experiments section) was demonstrated to have superior performance over regression~\cite{sinn2012predicting}. However, a great advantage of regression models is that they reveal which predictors have a significant effect on the response. Further, they provide a principled statistical framework for handling features and are highly interpretable. Our proposed EAM and AMM methods enjoy all the benefits of regression models but in addition allow for flexible modeling of nonlinear features through smooth functions.

\subsection{Kalman Filter Models}

Kalman filters~\cite{kalman1960new} and other time series models have been proposed for predicting bus arrival times~\cite{williams2003,vanajakshi2009travel,shalaby2004prediction,chien2003}. For the bus prediction problem the most common implementation of Kalman filter involves the assumption that travel time on a given route segment depends on a previously observed travel time at the same route segment~\cite{williams2003,shalaby2004prediction,chien2003}. However, \cite{vanajakshi2009travel} took a different approach and assumed that travel time on a given route segment depends on the travel time of a previous route segment. This approach resembles the SVM approach that was implemented for comparison purposes in the experiments section, see subsection below. All of the above methods rely on discretization of either time or space. \cite{williams2003} developed a seasonal autoregressive moving average process for short-term traffic forecasts. \cite{chien2003} treated the average travel time of tagged vehicles in a given time interval as the true value to predict the travel time in the next time period. In addition to using the travel time in current time interval,~\cite{shalaby2004prediction} also used the last three-day historical data of actual running times in the next time period to predict the next running time. The main limitation of a Kalman filter in the context of our data is the irregularity of observations. Large parts of the data contained time periods where no bus was observed and those time periods of missing data were generally different across different days. Therefore, a clear implementation strategy (e.g. time discretization) that covers all prediction scenarios would require some additional work. However, as noted in~\cite{altinkaya} Kalman filter give promising results on providing a dynamic travel time estimation. We note that both of our proposed EAM and AMM methods have a Kalman filter flavor as they include the last bus travel time as feature. 

\subsection{Artificial Neural Network Models}

Artificial Neural Network (ANN) models have gained recent popularity in predicting bus arrival times because of their ability to deal with complex and nonlinear relationships between variables~\cite{ramakrishna2006,chen2004dynamic,chien2002dynamic}. \cite{ramakrishna2006} developed an ANN model for prediction of bus travel times using GPS-based data and demonstrated superior performance over multiple linear regression. \cite{chen2004dynamic} developed an ANN model that further applied a dynamic Kalman filter algorithm to adjust predictions using bus location information. In order for the models in~\cite{chen2004dynamic,chien2002dynamic} to be practically implementable Automatic Passenger Count data need to be available in addition to the GPS data~\cite{altinkaya}. Additive Models in general share the ability of ANNs to flexibly deal with nonlinear relationships. However, they are further easily interpretable like regression models and do not suffer from slow learning process as reported for ANNs~\cite{altinkaya,hagan}. As implementation of ANN involves delicate setup of construction parameters (i.e., input variables, hidden layers, etc.) and none of the ANNs above were directly applicable to our setting we did not include ANNs in our experimental comparison. However, we did implement an SVM, discussed in the next subsection, which is a method that shares some functionalities with ANNs.

\subsection{Support Vector Regression Models}

SVM and Support Vector Regression (SVR) have demonstrated their success in time-series analysis and statistical learning~\cite{wu2004travel,bin@jits2006}. \cite{wu2004travel} compared their SVR algorithm to baseline predictors for prediction of travel time on highways and demonstrated superior performance. \cite{bin@jits2006} proposed SVM for travel time predictions and pointed out that unlike the traditional ANN, their method is not amenable to the overfitting problem. However, they also indicated that when SVM is applied for solving large problems the computation time becomes a problem.

\section{Conclusions}\label{sec:conclusion}

In this paper we discussed the problem of predicting travel times of public buses based on GPS data. We proposed Additive Models as a flexible and a statistically principled framework for model building. We modelled cumulative travel time as a sum of linear terms and smooth functions of predictor variables. We showed that by including a random intercept in the model we were able to correct for an interpolation error incurred when normalizing space-time trajectories onto a cumulative time scale. 
We demonstrated on a large real-world GPS data that our proposed Additive Models achieved superior performance as compared to other existing prediction methods.

\bibliographystyle{abbrv}
\bibliography{bibfile}

\end{document}